\documentclass[10pt,twocolumn,letterpaper]{article}

\usepackage{cvpr}
\usepackage{times}
\usepackage{epsfig}
\usepackage{graphicx}
\usepackage{amsmath}
\usepackage{amssymb}
\usepackage{algpseudocode}
\usepackage{algorithm}
\usepackage{caption}
\usepackage{subcaption}
\usepackage{multirow, booktabs}

\usepackage[pagebackref=true,breaklinks=true,letterpaper=true,colorlinks,bookmarks=false]{hyperref}

\cvprfinalcopy 

\def\cvprPaperID{13} 

\ifcvprfinal\fi
\begin{document}

\title{Attack Type Agnostic Perceptual Enhancement of Adversarial Images}

\author{Bilgin Aksoy and Alptekin Temizel\\
Middle East Technical University\\
Ankara, Turkey\\
{\tt\small \{bilgin.aksoy,atemizel\}@metu.edu.tr}
}

\maketitle

\begin{abstract}
Adversarial images are samples that are intentionally modified to deceive machine learning systems. They are widely used in applications such as CAPTHAs to help distinguish legitimate human users from bots. However, the noise introduced during the adversarial image generation process degrades the perceptual quality and introduces artificial colors; making it also difficult for humans to classify images and recognize objects. In this paper, we propose a method to enhance the perceptual quality of these adversarial images. The proposed method is attack type agnostic and could be used in association with the existing attacks in the literature. Our experiments show that the generated adversarial images have lower Euclidean distances to their originals while maintaining the same adversarial attack performance. Distances are reduced by 5.88\% to 41.27\% with an average reduction of 22\% over the different attack and network types.
\end{abstract}

\section{Introduction}
\textbf{C}ompletely \textbf{A}utomated \textbf{P}ublic \textbf{T}uring test to tell \textbf{C}omputers and \textbf{H}umans \textbf{A}part - CAPTCHA, is a commonly used method to validate human users. Image classification based tests are intentionally designed to make bots fail to classify images. Deep Neural Network (DNN) based methods \cite{goodfellow2013multi,stark2015captcha}, which have recently been proven to be successful in automated image classification, have been found to be useful to bypass CAPTCHA security process. However, these methods are vulnerable to specially generated adversarial examples \cite{Szegedy2013}, which can be used in CAPTCHAs and similar applications to make them more robust.

An adversarial attack perturbs the input image by adding a non-random, network and input specific noise, to make its automated classification difficult. This artificial noise also makes it more difficult for the legitimate users to classify the adversarial images especially when they are time limited \cite{gamaleddin}. So, two desired attributes of adversarial images are: (i) they should successfully fool the machine learning systems, (ii) they should introduce as little perceptual noise as possible so that they do not pose any additional challenge to the humans. In this paper, we propose a method for perceptual enhancement of adversarial images to make them closer to their noise-free originals and easier to process by humans. 

\section{Proposed Method}
\begin{figure}[t!]
	\centering
	\quad\quad\quad
	\begin{subfigure}{0.46\linewidth}
		\includegraphics[width=\linewidth]{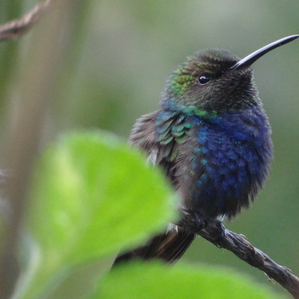}
		\caption{}
		\label{fig:org}
	\end{subfigure}
	\newline
	\begin{subfigure}{0.46\linewidth}
		\includegraphics[width=\linewidth]{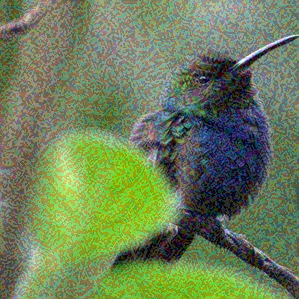}
		\caption{}
		\label{fig:fgsmbase}
	\end{subfigure}
	\begin{subfigure}{0.46\linewidth}
		\includegraphics[width=\linewidth]{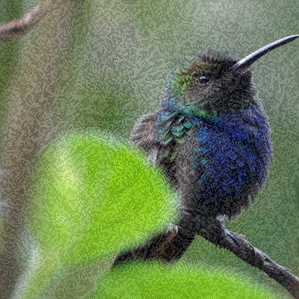}
		\caption{}
		\label{fig:fgsm0}
	\end{subfigure}
	\newline
	\begin{subfigure}{0.46\linewidth}
		\includegraphics[width=\linewidth]{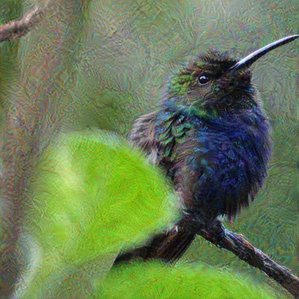}
		\caption{}
		\label{fig:cwbase}
	\end{subfigure}
	\begin{subfigure}{0.46\linewidth}
		\includegraphics[width=\linewidth]{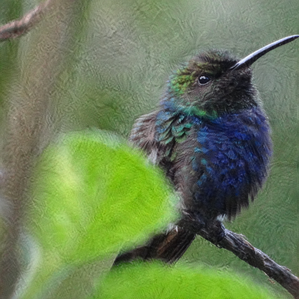}
		\caption{}
		\label{fig:cw0}
	\end{subfigure}
	\newline
	\begin{subfigure}{0.46\linewidth}
		\includegraphics[width=\linewidth]{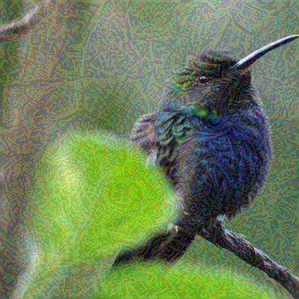}
		\caption{}
		\label{fig:mimbase}
	\end{subfigure}
	\begin{subfigure}{0.46\linewidth}
		\includegraphics[width=\linewidth]{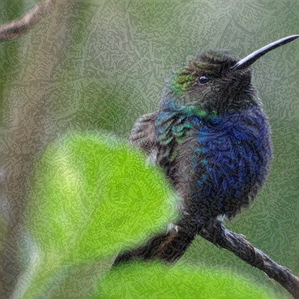}
		\caption{}
		\label{fig:mim0}
	\end{subfigure}
	\newline
	\caption{A sample image, its adversarial counterparts obtained using different attacks and with the proposed method. (a) Original image, (b) Baseline adversarial image (FGSM attack), (c) Adversarial image obtained with $\alpha=0$ (FGSM attack), (d) Baseline adversarial image (C\&W $L_2$ attack),  (e) Adversarial image obtained with $\alpha=0$  (C\&W $L_2$ attack),  (f) Baseline adversarial image (MIM attack),  (g) Adversarial image obtained with $\alpha=0$ (MIM attack)}
		\label{fig1}
\end{figure}

The inputs of conventional DNNs are RGB images and the attacks add noise to all three channels separately. Adding independent and different amounts of noise to these different channels results in artificial colors being introduced as shown in Fig.\ref{fig:fgsmbase}, \ref{fig:cwbase}, \ref{fig:mimbase}. In addition, as the attack modifies each pixel independently, it exhibits itself as a visually distractive colored snow-like high-frequency noise \cite{jpeg}. On the other hand, main distinguishing features (such as shape and texture) for an object class can be obtained from the luminance and adversarial noise added to the luminance channel is expected to be more detrimental to the network performance than the noise in the color channels. So, we claim that lower noise levels could be obtained by concentrating the attack on the luminance channel, which in effect is expected to reduce the distractive colored snow-like noise.\\
\begin{algorithm}
	\caption{Iteratively Finding the Minimum Adversarial Noise}
	\label{alg:pea}
	\begin{algorithmic}[1]
		\State Convert the original image $I^{R,G,B}$ into YUV: $I^{Y,U,V}$
		\State Initialize the best distance $L'_2$ to a high number 
		\While{Attack is successful}
		\State Run the attack to generate adversarial noise image $N^{R,G,B}$
		\State Convert $N^{R,G,B}$ into YUV: $N^{Y,U,V}$ 
		\State Scale the noise in U and V channels by a factor of $\alpha$, apply Gaussian smoothing $G$ to all noise channels and construct the adversarial image:
		\State $A^Y = I^Y + G(N^Y)$
		\State $A^U = I^U + G(\alpha \times N^U)$
		\State $A^V = I^V + G(\alpha \times N^V)$
		\State Convert $A^{Y,U,V}$ into RGB: $A^{R,G,B}$
		\State Calculate the new distance $L_2$ using $A^{R,G,B}$ and $I^{R,G,B}$
		\If{$L_2 < L'_2$ and attack is successful}
		\State Store the best attack:
		\State $A'^{R,G,B}$ = $A^{R,G,B}$ 
		\State Store the minimum $L_2$ value as the new minimum
		\State $L'_2$ = $L_2$
		\State Decrease the attack strength ($\epsilon$ for FGSM and MIM, maximum iteration for C\&W $L_2$)
		\Else{
			\Return $A'^{R,G,B}$}
		\EndIf
		\EndWhile
	\end{algorithmic}
\end{algorithm}
As conventional networks work with RGB images, the adversarial noise calculation inherently makes use of R, G and B channels. For the original image $I^{R,G,B}$, attack algorithm calculates the adversarial noise, $N^{R,G,B}$, separately for each channel. This noise is then added to the respective channels of the original image to obtain adversarial image $A^{R,G,B}$ as follows: $A^{R,G,B}=I^{R,G,B}+N^{R,G,B}$. In this work, we first convert the image and the adversarial noise into YUV domain and obtain $I^{Y,U,V}$ and $N^{Y,U,V}$ respectively. Then U and V coefficients of the noise, $N^{U}$ and $N^{V}$, are scaled by a factor $0\leq\alpha\leq1$. Assuming that the target object is closer to the center of the image, all the noise channels $N^{Y,U,V}$ are filtered with a 2D Gaussian kernel placed at the center of the image to gradually reduce the noise closer to the edges. The resulting noise is added in YUV color space: $A^{Y,U,V}=I^{Y,U,V}+N^{Y,U,V}$. Then the image $A^{Y,U,V}$ is converted back into RGB to allow processing in conventional networks. This process reduces the total amount of noise added to the original image and it might cause the adversarial attack to fail. Hence an iterative process is used as described in Alg.\ref{alg:pea} to find a stronger attack. Although a stronger attack will increase the noise, overall noise is lower due to the subsequent scaling of chrominance values and the use of Gaussian kernel.

\begin{algorithm}
	\caption{Iteratively Finding the Minimum Adversarial Noise}
	\label{alg:pea}
	\begin{algorithmic}[1]
		\State Convert the original image $I^{R,G,B}$ into YUV: $I^{Y,U,V}$
		\State Initialize the best distance $L'_2$ to a high number 
		\While{Attack is successful}
		\State Run the attack to generate adversarial noise image $N^{R,G,B}$
		\State Convert $N^{R,G,B}$ into YUV: $N^{Y,U,V}$ 
		\State Scale the noise in U and V channels by a factor of $\alpha$, apply Gaussian smoothing $G$ to all noise channels and construct the adversarial image:
		\State $A^Y = I^Y + G(N^Y)$
		\State $A^U = I^U + G(\alpha \times N^U)$
		\State $A^V = I^V + G(\alpha \times N^V)$
		\State Convert $A^{Y,U,V}$ into RGB: $A^{R,G,B}$
		\State Calculate the new distance $L_2$ using $A^{R,G,B}$ and $I^{R,G,B}$
		\If{$L_2 < L'_2$ and attack is successful}
		\State Store the best attack:
		\State $A'^{R,G,B}$ = $A^{R,G,B}$ 
		\State Store the minimum $L_2$ value as the new minimum
		\State $L'_2$ = $L_2$
		\State Decrease the attack strength ($\epsilon$ for FGSM and MIM, maximum iteration for C\&W $L_2$)
		\Else{
			\Return $A'^{R,G,B}$}
		\EndIf
		\EndWhile
	\end{algorithmic}
\end{algorithm}

\section{Dataset}
NIPS 2017: Adversarial Learning Development Set \cite{google2017} consist of 1000 images having 299x299 resolution. Each image corresponds to a different ImageNet 1000 category. Image pixels are scaled to the range $\left[0,1\right]$. All the images are used in the experiments and overall $L_2$ distances are calculated as the average throughout all the images.

\section{Experimental Setup}

$L_0$, $L_2$, and $L_\infty$ distances are commonly measures to quantify the perturbation added to the original image. $L_0$ distance counts the number of pixels which were altered during the adversarial process.  $L_\infty$ distance shows the maximum change due to the perturbation. Since  our method  aims perceptual enhancement, we calculate $L_2$ metric using all the channels (\ref{eq:l2}) in order to measure the total perturbation. In this equation, $I$ is the original image, $A$ is the adversarial image, $w$ is the width and, $h$ is the height of the image.  $L_2$ distance gives a better indication of the overall adversarial noise (high frequency noise which is distractive to human visual system) compared to $L_0$ and $L_\infty$.

\begin{equation}\label{eq:l2}
L_2 = \sqrt{\displaystyle\sum_{c}^{R,G,B} {\displaystyle\sum_{i=0}^{w} {\displaystyle\sum_{j=0}^{h} (I_{i,j}^c-A_{i,j}^c )^2 }} } 
\end{equation}

Fast Gradient Sign Method (FGSM) \cite{goodfellowexplaining}, Momentum Iterative Method (MIM) \cite{mim}  and Carlini\&Wagner $L_2$ (C\&W $L_2$) \cite{CarliniW16a}  attacks were used for experimental evaluation of the proposed method as they are well-known milestone attacks.

FGSM \cite{goodfellowexplaining} is a one-step gradient based approach which is designed to be fast. For a given image ${ \boldsymbol { I } }$ and corresponding target $y$, it calculates the gradient of the loss, $ \nabla _ { \boldsymbol { I } } J ( \boldsymbol { I } , y )$, generally cross-entropy, with respect to ${ \boldsymbol { I } }$ and multiplies negative of the gradient sign with a constant $\epsilon$  to generate the adversarial noise. This noise is then added to the image $I$ to obtain the adversarial example ${ \boldsymbol { A } }$ (\ref{eq:fgsm}). 

\begin{equation}\label{eq:fgsm}
{ \boldsymbol { A } }= { \boldsymbol { I } } - \epsilon \operatorname { sign } \left( \nabla _ { \boldsymbol { I } } J ( \boldsymbol { I } , y ) \right)
\end{equation}

MIM \cite{mim} is an iterative version of FGSM. It is designed to attain the minimum adversarial example in $T$ iterations. At each iteration, MIM updates the accumulated the gradient by using the current $L_1$ normalized gradient of loss, softmax cross-entropy, and previous accumulated gradient ${\boldsymbol{g_t}}$ multiplied by a decay factor $\mu$ (\ref{eq:g}). By this way, a momentum is used which makes the method more resilient to small humps, narrow valleys, and poor local extremities. Then the next adversarial example  ${ \boldsymbol { A_{t+1} } }$ is obtained by subtracting $L_2$ normalized ${\boldsymbol{g_{t+1}}} $ multiplied with a constant $\beta =  \frac{\epsilon}{T}$.

\begin{equation}\label{eq:g}
{\boldsymbol{g}}  _ { t + 1 } = \mu \cdot {\boldsymbol{g}}  _ { t } + \frac { \nabla _ { {\boldsymbol{I}} } J \left( {\boldsymbol{A}} _ { t } , y   \right) } { \left\| \nabla _ { {\boldsymbol{I}} } J \left( {\boldsymbol{A}}  _ { t } , y   \right) \right\| _ { 1 } }
\end{equation}

\begin{equation}\label{eq:mim}
{\boldsymbol{A}} _ { t + 1 } = {\boldsymbol{A}} _ { t } - \beta \cdot \frac { {\boldsymbol{g}} _ { t + 1 } } { \left\| {\boldsymbol{g}} _ { t + 1 } \right\| _ { 2 } }
\end{equation}

C\&W $L_2$ attack \cite{CarliniW16a} aims to find the lowest perturbation in $L_2$ distance metric, also in an iterative manner. At each iteration, the attack finds the perturbation $w$ for a given input image $ {\boldsymbol{I}}$ and target class $t$ by solving (\ref{eq:cw})

\begin{equation}\label{eq:cw}
\text{minimize}\left\| \frac { 1 } { 2 } (\tanh (w) + 1 )-{\boldsymbol{I}}\right\| _ {2}^{2} + c\cdot f\left(\frac{1}{2}(\tanh(w) + 1)\right)
\end{equation}
where $c$ is a constant and $f$ is defined as in (\ref{eq:f})
\begin{equation}\label{eq:f}
f \left( {\boldsymbol{A}} \right) = \max \left( \max \left\{ Z \left( {\boldsymbol{A}} \right) _ { i } : i \neq t \right\} - Z \left( {\boldsymbol{A}} \right) _ { t } , - \kappa \right)
\end{equation}
where $Z$ is the activation function and $\kappa$ is the confidence  parameter, (how confident the classifier should be that the generated adversarial image is a sample of the target class). In this work, we use a non-targeted setup so that $t$ is any incorrect class.

Cleverhans module \cite{papernot2016technical} was used for implementing the attacks. Each attack was trained in an untargeted setup and defended on three different pretrained network architectures: Inception v3 (IncV3) \cite{SzegedyVISW15}, InceptionResNet v2 (IncresV2) \cite{SzegedyIV16}, and ResNet50 v3 (Res50V3) \cite{HeZRS15}.

The experiments aim that all attacks are successful, i.e., the adversarial image generated by the attack network is misclassified by the defense network. To this end, $\epsilon$ parameter is used for FGSM and MIM attacks and iteration parameter is used for C\&W $L_2$ to find the minimum $L_2$ making the attack successful for each image.  The images are downscaled to 224x224 for Res50v3 and they are kept at their original resolution (229x299) for IncV3 and IncresV2. For all attack types, the Gaussian kernel size is set to match the size of the image and it has a standard deviation of 190.

For FGSM attack, $\epsilon$ parameter is selected as 10.0 at the first iteration and decreased by 0.025 until the minimum $\epsilon$ which makes the defense network misclassify the adversarial image is obtained. If the adversarial attack fails at the first iteration then $\epsilon$ is increased by 5.0 and if it is successful then decreased by 0.025 until the minimum $\epsilon$ that makes the network misclassify the input is obtained. 

C\&W $L_2$ attack is initialized by setting confidence parameter to zero. Then the iteration parameter is increased, as long as the attack is successful, to find the minimum $L_2$ distance.

For MIM attack, $\epsilon$ parameter is selected as 0.018 for the first iteration and decreased by 0.001 until the minimum $L_2$ distance is obtained.

\section{Experimental Results}
The results are shown in Table \ref{tab:1} for different $\alpha$ values where baseline refers to the original unmodified attack.  Note that the case where  $\alpha$ is 1 still has an effect of reducing the noise due to the Gaussian smoothing. When $\alpha$ is 0, no noise is added to the color channels.

Fig. \ref{fig1} shows baseline adversarial images and the images obtained with the proposed method for FGSM, C\&W $L_2$ and MIM attacks. 

\begin{figure}[t!]
	\centering
	\includegraphics[width=1\linewidth]{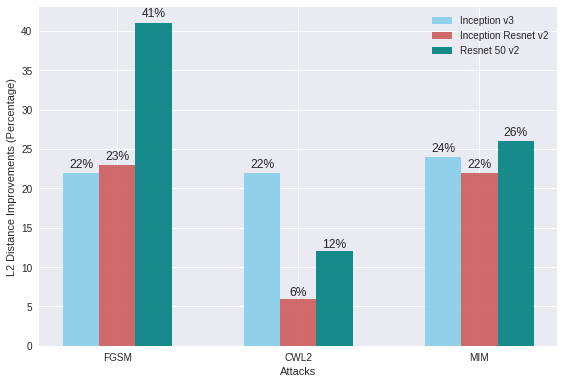}
	\caption{$L_2$ distance improvements with respect to base attack for different attack types and networks}
	\label{fig:imp}
\end{figure}

\begin{figure}[t!]
	\centering
	\quad\quad\quad
	\begin{subfigure}{0.46\linewidth}
		\includegraphics[width=\linewidth]{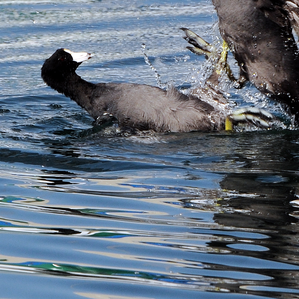}
		\caption{}
		\label{fig:2org}
	\end{subfigure}
	\newline
	\begin{subfigure}{0.46\linewidth}
		\includegraphics[width=\linewidth]{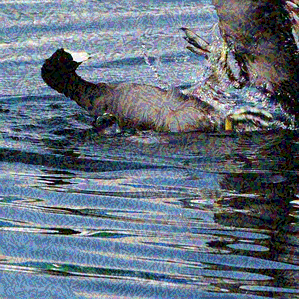}
		\caption{}
		\label{fig:2fgsmbase}
	\end{subfigure}
	\begin{subfigure}{0.46\linewidth}
		\includegraphics[width=\linewidth]{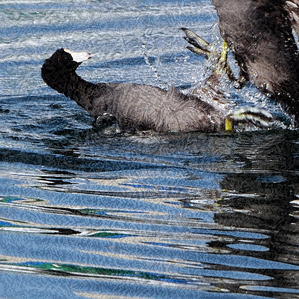}
		\caption{}
		\label{fig:2fgsm0}
	\end{subfigure}
	\newline
	\begin{subfigure}{0.46\linewidth}
		\includegraphics[width=\linewidth]{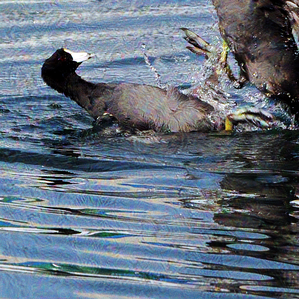}
		\caption{}
		\label{fig:2cwbase}
	\end{subfigure}
	\begin{subfigure}{0.46\linewidth}
		\includegraphics[width=\linewidth]{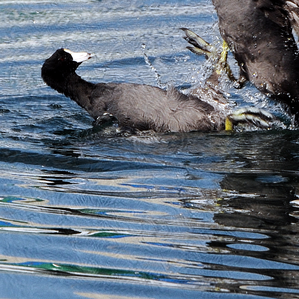}
		\caption{}
		\label{fig:2cw0}
	\end{subfigure}
	\newline
	\begin{subfigure}{0.46\linewidth}
		\includegraphics[width=\linewidth]{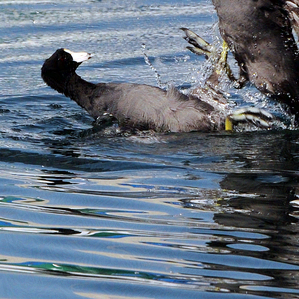}
		\caption{}
		\label{fig:2mimbase}
	\end{subfigure}
	\begin{subfigure}{0.46\linewidth}
		\includegraphics[width=\linewidth]{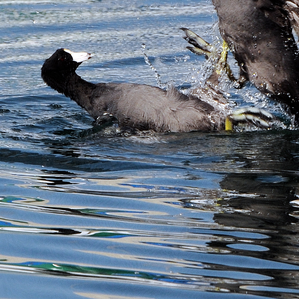}
		\caption{}
		\label{fig:2mim0}
	\end{subfigure}
	\newline
	\caption{A sample image, its adversarial counterparts obtained using different attacks and with the proposed method. (a) Original image, (b) Baseline adversarial image (FGSM attack), (c) Adversarial image obtained with $\alpha=0$ (FGSM attack), (d) Baseline adversarial image (C\&W $L_2$ attack),  (e) Adversarial image obtained with $\alpha=0$  (C\&W $L_2$ attack),  (f) Baseline adversarial image (MIM attack),  (g) Adversarial image obtained with $\alpha=0$ (MIM attack)}
	\label{fig2}
\end{figure}

Fig. \ref{fig:imp} shows $L_2$ distance improvements as percentage of the baseline attacks.  The largest improvement is obtained for FGSM using Res50V3 where it is improved by 41.27\% and smallest improvement is 5.88\% for C\&W $L_2$ using IncresV3. On average 22\% improvement is achieved considering all attack and network types. Another sample image and its adversarial counterparts are shown in Fig.\ref{fig2}.

\begin{table*}[t!]
	\centering
	\caption{$L_2$ distances for different attacks and different networks using various $\alpha$ values}
		\begin{tabular}{|l|c|c|c|c|c|c|c|c|c|}
			\toprule
			\multirow{2}{*}{Method}			&\multicolumn{3}{c|}{FGSM}	&\multicolumn{3}{c|}{C\&W $L_2$} &\multicolumn{3}{c|}{MIM} \\
			\cmidrule{2-10}
			& IncV3 & IncresV2  & Res50V3	& IncV3 & IncresV2  & Res50V3& IncV3 & IncresV2  & Res50V3	\\ 
			\midrule
			Baseline        & 4.3029 &40.02 & 40.78&0.2285&0.3484    &8.0870 	&0.6012&0.8290      &0.3859\\ 
			$\alpha=1$      & \textbf{3.3605}& 36.98   &30.19 & 0.1996 & 0.3478 &   7.1845 &0.5932&0.7492&0.3190\\ 
			$\alpha=0.8$  & 3.7134 &  35.17  & 28.14   &0.1888      &  0.3382  & 7.1365& 0.5411 & 0.6857 & 0.3052 \\ 
			$\alpha=0.6$  &  4.3884& 33.12   & 25.54  &0.181   &  \textbf{0.3279}  &  \textbf{7.1373}& 0.4995 &0.6662&  0.2947 \\ 
			$\alpha=0.4$  & 4.1897 &  31.72   & 24.04  &\textbf{0.1782}  & 0.3306   & 7.3007& 0.4712 &0.6474 &  0.2877\\ 
			$\alpha=0.2$  & 4.6203& \textbf{30.71}   &\textbf{23.95}  &0.1795  & 0.3306   & 7.3007 &  \textbf{0.4589} &\textbf{0.6431} &  \textbf{0.2846} \\ 
			$\alpha=0$    &  6.3628& 35.72  & 25.87 & 0.1863  &  0.3425   &   7.4638 &0.4638 & 0.6486 &   0.2854 \\ 
			\bottomrule
		\end{tabular}
\label{tab:1}
\end{table*}

\section{Discussion} When we reduce the noise in U and V bands, the adversarial images look perceptually better. However, in order to achieve 100\% attack accuracy, stronger attacks, which increase the noise in Y, are needed as a trade-off. However,  as can be seen in Table \ref{tab:1}, lower $L_2$ distances can still be obtained for all attack types and for all networks. It has to be noted that the $\alpha$ value giving the best result is different for each attack. For FGSM, $\alpha = 0.2$ gives the best results for IncresV2 and Res50V3 while $\alpha = 1$ is the best for IncV3. For C\&W $L_2$, $\alpha = 0.6$ gives the best results for IncresV2 and Res50V3. While $\alpha = 0.8$ is the best for IncV3, performance difference with $\alpha = 0.6$ is relatively small and it can be said that, in practice, $\alpha = 0.6$ can be used for all network types in question. For MIM,  $\alpha = 0.2$ gives the best results for all different types of networks in question.

The results show that the proposed method works independent of the attack type and the network model and reduces the $L_2$ distances. Even though C\&W $L_2$ and MIM attacks are optimized to minimize $L_2$ distance by design, our method results in still lower $L_2$ values. While this might sound contradictory, it has to be noted that due to the nature of the networks, this optimization is done on RGB values in the original attacks and might not be optimal when YUV domain is considered.  The proposed method reduces the noise in U and V channels which is compensated by increasing the noise in Y channel. This strategy reduces the amount of perceptible color noise as well as reducing the total noise as indicated by $L_2$ distances calculated using RGB channels.

Since C\&W $L_2$ and MIM generate adversarial noise in iterative manner, both are able to produce lower $L_2$ distance than FGSM. C\&W $L_2$ attack  achieved the best $L_2$ distances except using ResNet50v3 as the attack network. For this network, MIM attack achieved the best $L_2$ distance.

\section{Conclusion} We proposed an attack and network type agnostic perceptual enhancement method  by converting the adversarial noise to YUV color space and reducing the chrominance noise and applying Gaussian smoothing to the adversarial noise. The adversarial images are not only perceptually better but also have lower $L_2$ distances to the original images. Conventional networks are trained using images in RGB color space and inherently, the optimization is done in this color space. In the future, these networks could be trained using images in YUV color space. Then using these networks, attacks could be done intrinsically in YUV space.

The proposed method assumes that the object is located near the center of the image and Gaussian kernel is positioned at the center of the image. However the object could be off-center or could be located in a different position which might invalidate this assumption. In the future, class activation maps \cite{zhou2016learning}, which could be obtained directly through the attack network, can be used to estimate the center position of the object. This would allow positioning the Gaussian kernel to overlap better with the object position.

{\small
\bibliographystyle{ieee_fullname}
\bibliography{perceptualbib}

\begin{thebibliography}{10}\itemsep=-1pt

\bibitem{jpeg}
Ayse~Elvan Aydemir, Alptekin Temizel, and Tugba~Taskaya Temizel.
\newblock The effects of {JPEG} and {JPEG2000} compression on attacks using
  adversarial examples.
\newblock {\em arXiv preprint arXiv:1803.10418}, 2018.

\bibitem{CarliniW16a}
Nicholas Carlini and David Wagner.
\newblock Towards evaluating the robustness of neural networks.
\newblock {\em {IEEE} Symposium on Security and Privacy {(SP)}}, pages 39--57,
  2017.

\bibitem{mim}
Yinpeng Dong, Fangzhou Liao, Tianyu Pang, Hang Su, Jun Zhu, Xiaolin Hu, and
  Jianguo Li.
\newblock Boosting adversarial attacks with momentum.
\newblock {\em A Proceedings of the {IEEE} Conference on Computer Vision and
  Pattern Recognition}, pages 9185--9193, 2016.

\bibitem{gamaleddin}
Gamaleldin~F. Elsayed, Shreya Shankar, Brian Cheung, Nicolas Papernot, Alex
  Kurakin, Ian Goodfellow, and Jascha Sohl-Dickstein.
\newblock Adversarial examples that fool both computer vision and time-limited
  humans.
\newblock {\em Advances in Neural Information Processing Systems}, pages
  3914--3924, 2018.

\bibitem{goodfellow2013multi}
Ian~J. Goodfellow, Yaroslav Bulatov, Julian Ibarz, Sacha Arnoud, and Vinay
  Shet.
\newblock Multi-digit number recognition from street view imagery using deep
  convolutional neural networks.
\newblock {\em International Conference on Learning Representations}, 2014.

\bibitem{Szegedy2013}
Ian~J. Goodfellow, Yaroslav Bulatov, Julian Ibarz, Sacha Arnoud, and Vinay
  Shet.
\newblock Multi-digit number recognition from street view imagery using deep
  convolutional neural networks.
\newblock {\em International Conference on Learning Representations}, 2014.

\bibitem{goodfellowexplaining}
Ian~J. Goodfellow, Jonathon Shlens, and Christian Szegedy.
\newblock Explaining and harnessing adversarial examples.
\newblock {\em arXiv preprint arXiv:1412.6572}, 2014.

\bibitem{HeZRS15}
Kaiming He, Xiangyu Zhang, Shaoqing Ren, and Jian Sun.
\newblock Deep residual learning for image recognition.
\newblock {\em {IEEE} Conference on Computer Vision and Pattern Recognition},
  pages 770--778, 2016.

\bibitem{google2017}
Kaggle.
\newblock Nips 2017: Adversarial learning development set, Jul 2017.

\bibitem{papernot2016technical}
Nicolas Papernot, Fartash Faghri, Nicholas Carlini, Ian Goodfellow, Reuben
  Feinman, Alexey Kurakin, Cihang Xie, Yash Sharma, Tom Brown, Aurko Roy,
  Alexander Matyasko, Vahid Behzadan, Karen Hambardzumyan, Zhishuai Zhang,
  Yi-Lin Juang, Zhi Li, Ryan Sheatsley, Abhibhav Garg, Jonathan Uesato, Willi
  Gierke, Yinpeng Dong, David Berthelot, Paul Hendricks, Jonas Rauber, Rujun
  Long, and Patrick McDaniel.
\newblock Technical report on the cleverhans v2. 1.0 adversarial examples
  library.
\newblock {\em arXiv preprint arXiv:1610.00768}, 2016.

\bibitem{stark2015captcha}
Fabian Stark, Caner Hazırbaş, Rudolph Triebel, and Daniel Cremers.
\newblock Captcha recognition with active deep learning.
\newblock {\em GCPR Workshop on New Challenges in Neural Computation}, 10,
  2015.

\bibitem{SzegedyIV16}
Christian Szegedy, Sergey Ioffe, Vincent Vanhoucke, and Alex Alemi.
\newblock Inception-v4, inception-resnet and the impact of residual connections
  on learning.
\newblock {\em Proceedings of the Thirty-First {AAAI} Conference on Artificial
  Intelligence}, pages 4278--4284, 2017.

\bibitem{SzegedyVISW15}
Christian Szegedy, Vincent Vanhoucke, Sergey Ioffe, Jonathon Shlens, and
  Zbigniew Wojna.
\newblock Rethinking the inception architecture for computer vision.
\newblock {\em Proceedings of the {IEEE} Conference on Computer Vision and
  Pattern Recognition}, pages 2818--2826, 2016.

\bibitem{zhou2016learning}
Bolei Zhou, Aditya Khosla, Agata Lapedriza, Aude Oliva, and Antonio Torralba.
\newblock Learning deep features for discriminative localization.
\newblock {\em {IEEE} Conference on Computer Vision and Pattern Recognition},
  pages 2921--2929, 2016.

\end{thebibliography}
}

\end{document}